# Development of a peristaltic micropump for bio-medical applications based on mini LIPCA

My Pham[1,3], Thanh Tung Nguyen[2,3], Nam Seo Goo[1,3]

[1]Department of Advanced Technology Fusion

[2]Department of Aerospace Engineering

[3]Biomemitics & Intelligent Microsystem Laboratory, Artificial Muscle Research Center

Konkuk University, 1 Hwayang-dong, Gwangjin-gu, Seoul 143-701, South Korea

Email: nsgoo@konkuk.ac.kr

**Abstract**

This paper presents the design, fabrication, and experimental characterization of a peristaltic micropump. The micropump is composed of two layers fabricated from polydimethylsiloxane (PDMS) material. The first layer has a rectangular channel and two valve seals. Three rectangular mini lightweight piezo-composite actuators are integrated in the second layer, and used as actuation parts. Two layers are bonded, and covered by two polymethyl methacrylate (PMMA) plates, which help increase the stiffness of the micropump. A maximum flow rate of 900 μl/min and a maximum backpressure of 1.8 kPa are recorded when water is used as pump liquid. We measured the power consumption of the micropump. The micropump is found to be a promising candidate for bio-medical application due to its bio-compatibility, portability, bidirectionality, and simple effective design.

## 1 Introduction

In a peristaltic pump, sequential actuations of actuating diaphragms in a desired fashion can generate fluid flow rates in a controlled direction. The main advantage of a peristaltic pump is that no valves or other internal parts ever touch running fluid. Due to their cleanliness, peristaltic pumps have found many applications in the pharmaceutical, chemical, and food industries. Besides this, the action of a peristaltic pump is very gentle, which is important when the fluid is easily damaged. Therefore, peristaltic pumps can be used in medical applications, one of which is moving the blood through the body during open heart surgery. The other type pumps would risk destroying the blood cells.

Previously a number of peristaltic micropumps have been reported with various driving mechanisms such as piezoelectric, thermopneumatic, electromagnetic, pneumatic, and electrostatic actuations [1,2]. Thermopneumatic actuator was reported in [3,4]. However, overheating in thermopnematic actuators can affect pumping biomaterials. Furthermore, long response time and high power consumption are other disadvantages of this actuator. PDMS peristaltic micropump with pneumatic actuation was experimentally studied and modeled in [5]. Their sizes were quite large, and thus they seemed not to be adequate for portable use. Design and simulation of a novel electrostatic peristaltic micropump was reported in [6]. A low output force and small stroke were found to be disadvantages of the electrostatic actuations. Piezoelectric actuations are the most popular ones applied for micropumps, because they normally have a short response time and a high output force [7]. However, high operational voltages and a small volume stroke are considered as their disadvantages.

In this paper, a low-cost, simple-design, relatively low operational voltage, and high performance peristaltic micropump was designed and fabricated with PDMS material, which has been widely used to fabricate microfluidic systems, due to its compatibility, transparent character, desired mechanical properties, and easy processing using micro molding techniques [8].

High volume stroke and relatively low voltage actuation diaphragm has been fabricated by using mini LIPCAs (lightweight piezo-composite actuators). Designed LIPCAs bonded with a thin layer of PDMS material was adopted instead of a conventional unimorph-type element. LIPCA, first introduced as a compact curved actuator device by Yoon et al. [9], represents a new class of piezoceramic-based actuator capable of generating significant displacement and forces in response to input voltages. A valveless micropump with LIPCA actuating diaphragm was previously introduced by the authors in [10]. With this actuating diaphragm, the piezoceramic wafer and electrode were effectively insulated from the environment.

To evaluate the performance of the micropump, the diaphragm displacement, the pump flow rate and head pressure were measured as function of applied voltages and frequencies. The testing results indicate that our type of actuator diaphragm offers a significant advantage over the conventional ones; increased displacement- several times that of a geometrically equivalent unimorph and bimorph-type actuators. From the experimental results, the present micropump could generate a flow rate up to 900μl/min with only 160 Vp-p (voltage peak to peak) applied voltage and 60 Hz frequency, higher than those of the previously developed peristaltic micropumps [11-15]. Finally, we have investigated the power consumption aspects of the developed micropump.

## 2. Design and fabrication of mini LIPCA

A LIPCA is a kind of multilayer composite actuator that is typically composed of glass-epoxy layers with a low modulus and a high coefficient of thermal expansion CTE, carbon-epoxy layers with a high modulus and a low CTE, and a layer consisting of a PZT ceramic wafer [9]. Figure 1 shows the lay-up structure and shape of the designed mini LIPCA. After the hand lay-up, the stacked laminate was vacuum-bagged and cured at 177 oC for 2 hours in an autoclave. For the peristaltic micropump three identical mini LIPCAs were fabricated. It is noted that 0.1 mm thick PZT was used for the fabrication of the LIPCAs in order to reduce the operational voltage of the LIPCAs. Figure 2 shows the photograph of the fabricated LIPCAs.

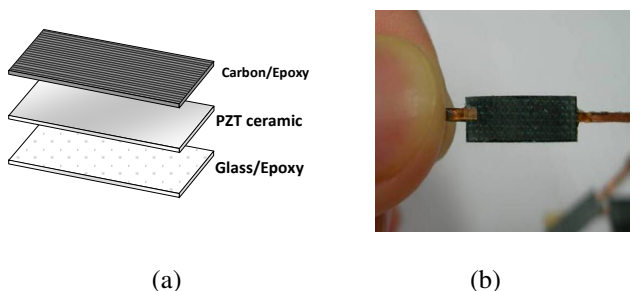

(a)             (b)

Figure 1 (a) Structure and shape of the designed mini LIPCA (b) Photograph of the fabricated mini LIPCAs

The deflection of the fabricated mini LIPCAs was measured in the simply support condition by using non-contact laser sensor (KEYENCE LK–G80). To excite the actuator we set the power supply (TD–2, Face International) at 1.0 Hz and up to 160 Vp–p. In addition, the commercial finite element software (MSC/NASTRAN) together with the thermal analogy technique was used to predict the deflection of the LIPCAs. Table 1 shows the basic properties of the constituent materials of the LIPCAs. Figure 2 shows the experimental and numerical results. Figure 2 shows that the actuating displacements increase at high applied voltages, probably due to the material nonlinearity of PZT material.

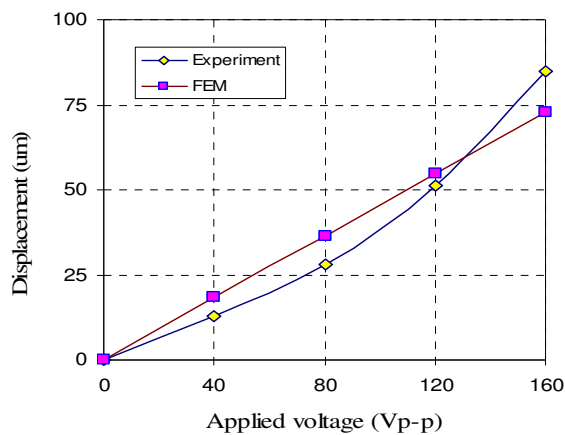

Figure 2 Comparison of the numerical and experimental results for the deflection of the mini LIPCAs.

Table 1 Basic properties of actuator materials

| Properties | | PZT3203 HD | Carbon/ Epoxy | Glass/ Epoxy |
|---|---|---|---|---|
| Modulus ($GPa$) | $E_1$ | 62 | 231.2 | 21.7 |
| | $E_2$ | 62 | 7.2 | 21.7 |
| | $G_{12}$ | 23.7 | 4.3 | 3.99 |
| Poison's ratio | $\nu_{12}$ | 0.31 | 0.29 | 0.13 |
| CTE ($10^{-6}/°K$) | $\alpha_1$ | 3.5 | –1.58 | 14.2 |
| | $\alpha_2$ | 3.5 | 32.2 | 14.2 |
| (*) $d_{31}$ ($10^{-12}$m/V) | | -320 | - | - |

(*) Piezoelectric strain coefficient

## 3. Micropump design and working principle

Figure 3 describes the stack-type structure of the developed peristaltic micropump. The micropump is composed of four layers of the same size. The first layer is a PMMA plate which has an opening at the center; the second layer is the LIPCA/PDMS diaphragm. The diaphragm has three mini LIPCAs, which were integrated in a PDMS layer; the third layer, made from PDMS, has two valve seals fabricated from SU-8 and two channels for inlet and outlet connected to the inlet and outlet piles. The pump chamber is defined by the space between the second layer and the third layer; the fourth layer fabricated form PMMA is used to increase stiffness of the micropump. The presence of the valve seals enables complete seals when the mini LIPCA deflects downward. All layers have four holes at the corners to ensure each stack is fixed with every other stack using four screws.

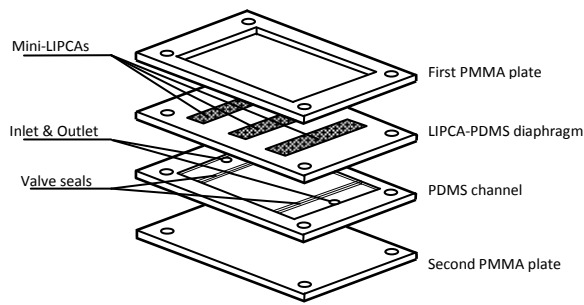

Figure 3 Structure of the micropump with four stacked layers.

Figure 4(a) describes the working principle of the peristaltic micropump. When driving signals as shown in figure 4(b) is applied to the three mini LIPCAs, these LIPCAs will deflect up and down according to signs of driving signals. This will induce a peristaltic motion of the diaphragm, which in turn conveys the pump liquid in the pump chamber. Because at least one of the valve seals is always sealed by the corresponding LIPCA, there is no flow in reverse direction, which enables a net flow rate in a specific direction. Direction of the pump flow can be chosen easily by controlling the driving signal.

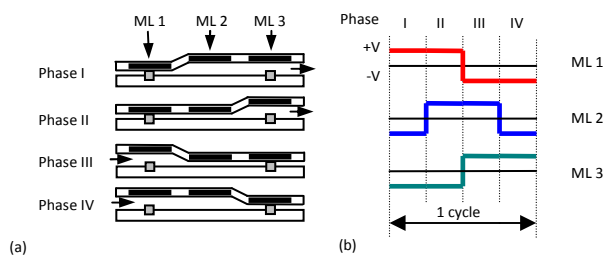

Figure 4 Operating principle of the micropump: (a) peristaltic motion of mini LIPCAs; (b) driving signal for three mini LIPCAs.

### 4. Fabrication process

Figure 5 describes the fabrication process of the LIPCA/PDMS diaphragm and PDMS channel depicted in Figure 3. To fabricate the LIPCA/PDMS diaphragm, a 100 µm-thick PDMS pre-polymer mixture was spin-coated on a silicon wafer and thermally cured (a). Next, an approximately 300 µm-thick PDMS pre-polymer mixture was created on the cured PDMS layer using spin coating process. Three fabricated mini LIPCAs were placed, aligned on the uncured PDMS layer, and pushed them so that they could contact the 100 µm-thick PDMS layer. This PDMS layer was then cured at 600 C for 2 hours (b). Another layer of 100 µm-thick PDMS pre-polymer mixture was afterward spin-coated on the LIPCA layer, and cured at 600 C for 2 hours (c). After cured the diaphragm was cut and punched into the shape, peeled off from the silicon wafer, and ready to be packaged with other layers (see figure 6(a)). To fabricate the second PDMS channel layer a 50 µm-thick SU-8 layer was created on a silicon wafer by using spin coating process (e). The layer was then patterned by using a standard photolithography process to create two mortises (f). PDMS pre-polymer mixture was then poured on the mold, and cured at 600 C for 2 hours (g). The PDMS layer was then carefully peeled off from the silicon wafer (h), aligned, and bonded with the LIPCA/PDMS diaphragm in step (c) (d).

The PMMA parts including the first and the fourth layer were fabricated using conventional milling machining. The four layers of the micropump were stacked and fixed by using four screws. Finally, plastic piles were connected to the inlet and outlet holes, resulting in a complete micropump as shown in figure 6(b). The overall size of the micropump is 20×16×4 mm3.

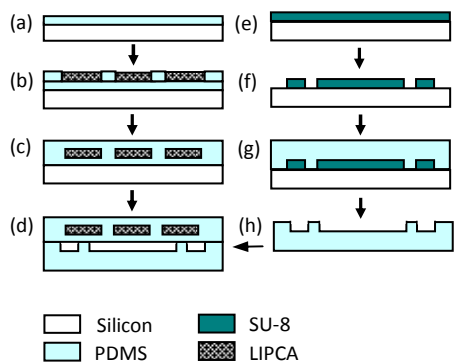

Figure 5 The fabrication process of the micropump: (a) PDMS curing; (b) bonding LIPCAs with the PDMS layer; (c) covering LIPCA with a PDMS layer; (e)&(f) patterning SU-8 mold; (g)&(h) molding and peeling PDMS; (d) bonding PDMS layers.

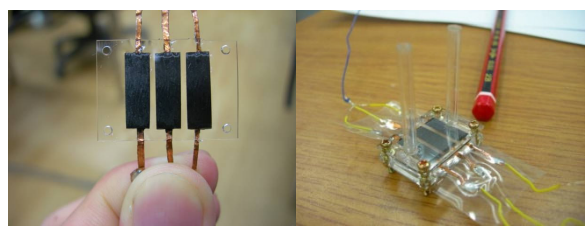

Figure 6 Photography of (a) the fabricated LIPCA diaphragm and (b) the fabricated micropump

### 5. Experiments and results

The pump characteristics were measured with water as the pump liquid. The water flow rate was measured under various applied voltages, frequencies, and backpressures. The flow rate was calculated from the weight of pumped water in the reservoir and the corresponding experimental period. Figure 7 shows the experiment setup for the flow rate tests. To generate driving voltage signal as in Figure 4(b), we made a power supply producing three output voltages with different phase angles and magnitude.

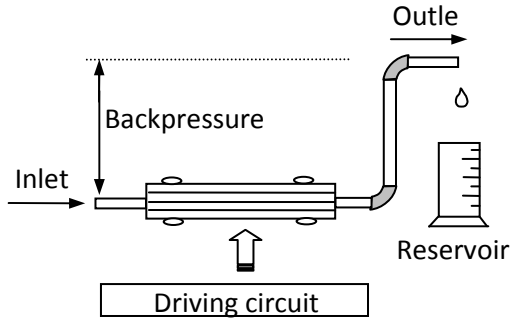

Figure 7 Experiment setup for the micropump flow rate tests.

Power consumption is also measured with respect to various operation voltages and frequencies. A measurement system including the micropump, three external resistors, a power supply, and an oscilloscope (Tektronix TDS 2024) was set up as in Figure 8. Each LIPCA was attached to a 1KΩ external resistor in series and voltage over of each external resistor was monitored to determine the current with a relation, $i_k(t) = V_k^E(t)/R_E, k = 1, 2, 3$

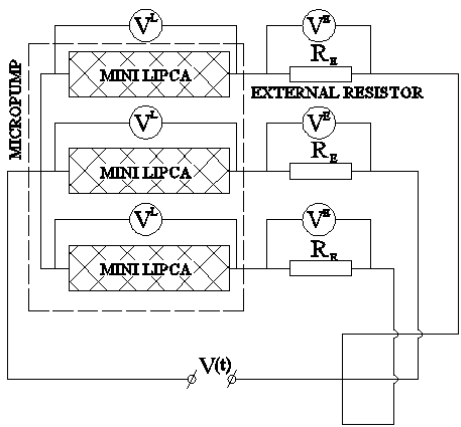

Figure 8 Experimental setup for the measurement of the power consumption

A sinusoidal applied voltage $V(t)$ was imposed at the each LIPCA and the phase difference of the applied voltage was 120 degrees. We changed the applied voltage from 20 to 160 Vp-p and driven frequencies ranging from 10 to 160 Hz. While the micropump was operated with water fluid, the voltage over each LIPCA $V_k^L(t)$ and the voltage over each resistor $V_k^E(t)$ were recorded into the memory of an oscilloscope. The power consumption for the micropump could be calculated by summation of the power consumption of each LIPCA, written as the following formula:

$$P = \sum_{k=1}^{3} \left[ \frac{1}{T} \int_0^T V_k^L(t) i_k(t) dt \right] \quad (1)$$

where T is a period.

## 6. Results and discussions

Figure 9(a) presents the flow rate versus driven frequency of the micropump measured at 80 Vp-p (Volts peak to peak). As seen in figure 9(a), the flow rate reaches a maximum at around 60 Hz and reduces rapidly as driven frequency becomes higher. Figure 9(b) presents the flow rates with respect to applied voltage of the micropump measured at 10 Hz and 60 Hz. Because the thickness of PZT is 0.1mm, we confined the applied voltage to 160 Vp-p to avoid the domain switching phenomenon. Whereas the flow rate is almost negligible below 40 Vp-p, the flow rate gets higher as the applied voltage becomes larger above 40 Vp-p. Figure 9(c) shows the maximum backpressure recorded at a frequency of 60 Hz and different applied voltages. In figure 9(d), the flow rate versus the backpressure curve recorded at 160 Vp-p and 60 Hz is given. Whereas the micropump exhibits a maximum backpressure of 1.8 kPa at zero flow rate, the flow rate reached the maximum value of 900 μl/min at zero backpressure.

Figure 10 shows the power consumption of the present micropump with respect to driven frequency, applied voltage and flow rate. From Fig. 10 (a), we found that the maximum power consumption is at a driven frequency of 60 Hz from 20 to 160 Vp-p. Fig. 10 (b) shows that the power consumption increases rapidly as the flow rate increases. With a power consumption of as small as 45 mW, the micropump has achieved a flow rate up to 900 μl at 160 Vp-p and 60 Hz. This is a promising result being compared with the other micropump that used the other actuator configurations, for instance, micropump using thermo-pneumatic actuation by Sim et al [16] and Grosjean and Tai [17]. One of important factors for the efficient micropump is power consumption. Small power consumption is necessary for bio-medical applications. We used 0.1mm thick PZT wafers to reduce the applied voltage and made an efficient design to reduce the power consumption.

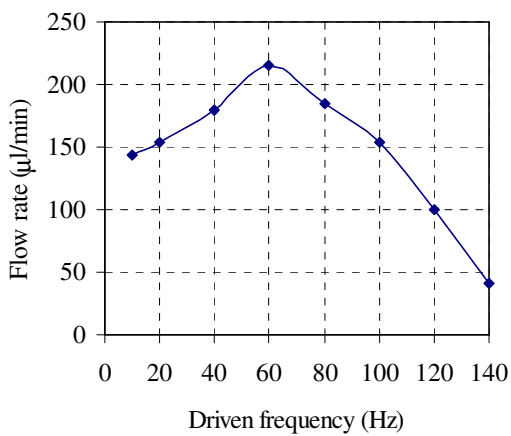

(a) Measured flow rate versus the driven frequencies at 80 Vp-p

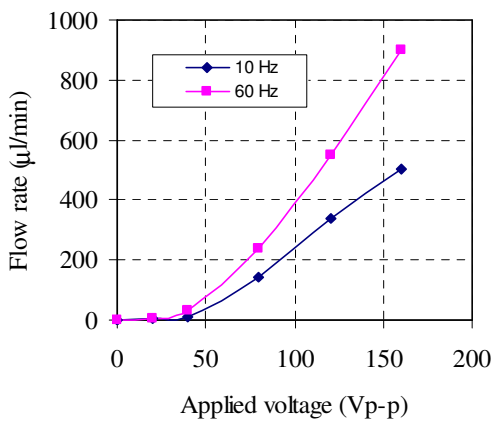

(b) Measured flow rate versus the applied voltage at 10 and 60 Hz

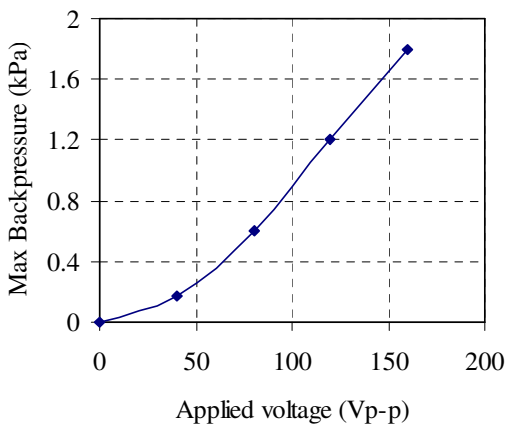

(c) Maximum backpressure versus applied voltage at a frequency of 60 Hz

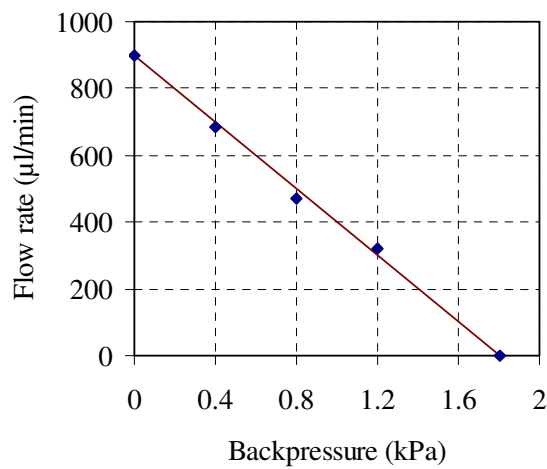

(d) Measured flow rate versus back pressure at an applied voltage of 160 Vp-p and a driven frequency of 60 Hz

Figure 9 Pumping performance measurement results

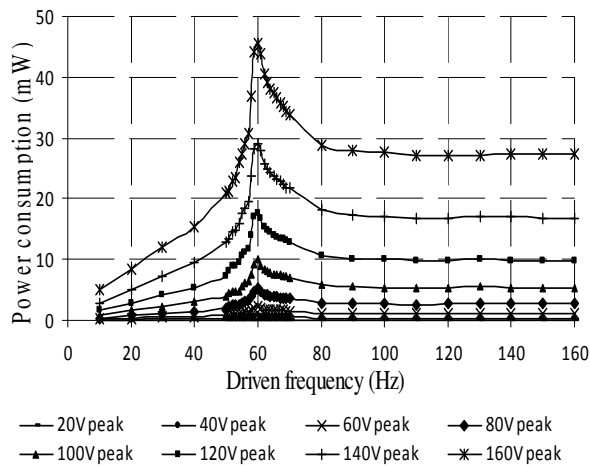

(a) Power consumption vs. driven frequency

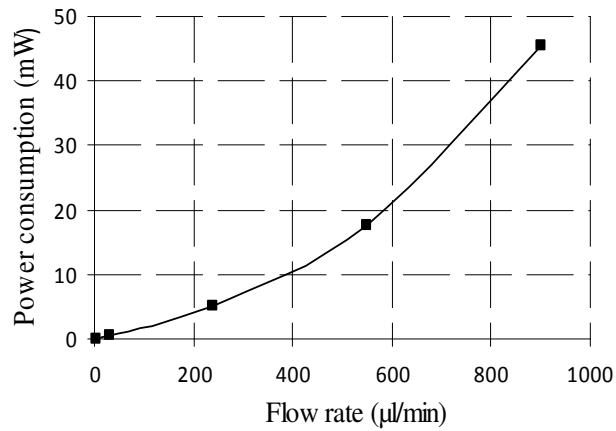

(b) Flow rate vs. power consumption at an applied voltage of 160 Vp-p and a driven frequency of 60 Hz

Figure 10 Power consumption measurement results.

## 7. Summary

In this paper we present the design, fabrication, and experimentally characterization of a peristaltic micropump. Three rectangular mini LIPCAs are fabricated, and used as actuation sections. Their deflections were numerically predicted and measured with a non-contact laser sensor. Two PDMS layers of the pump were fabricated by using micro molding. They are sandwiched between two PMMA plates, which are fixed by four screws to complete the micropump. The flow rate of the micropump is measured at various working conditions. The micropump reaches a maximum water flow rate of 900 μl/min, and a maximum backpressure of 1.8 kPa. We also considered the power consumption of the developed micropump. The proposed micropump offers many advantages including a comparatively low applied voltage, simple and effective design providing for ease of manufacturing, and bio-compatibility.

## Acknowledgement

This work is supported by a Korea Research Foundation Grant (KRF-2006-331-D00085 and KRF–2006-005–J03302). The authors are grateful for the financial support.